\documentclass[10pt,twocolumn,letterpaper]{article}

\usepackage{cvpr}
\usepackage{times}
\usepackage[utf8]{inputenc} 
\usepackage[T1]{fontenc}    
\usepackage{epsfig}
\usepackage{graphicx}
\usepackage{amsmath}
\usepackage{amssymb}
\usepackage{url}
\usepackage{flushend}
\usepackage{comment}
\usepackage{adjustbox}
\usepackage{authblk}
\usepackage[noend]{algpseudocode}
\usepackage{subcaption}
\usepackage[ruled]{algorithm2e}

\usepackage[breaklinks=true,bookmarks=false]{hyperref}

\cvprfinalcopy 

\def\cvprPaperID{****} 

\usepackage{amsmath,amsfonts,bm}

\def\eqref#1{equation~\ref{#1}}

\def\1{\bm{1}}

\DeclareMathAlphabet{\mathsfit}{\encodingdefault}{\sfdefault}{m}{sl}
\SetMathAlphabet{\mathsfit}{bold}{\encodingdefault}{\sfdefault}{bx}{n}

\begin{document}

\title{Semi-Weakly Supervised Object Detection by Sampling\\ Pseudo Ground-Truth Boxes}

\author[$1$]{Akhil Meethal}
\author[$1$]{Marco Pedersoli}
\author[$2$]{Zhongwen Zhu}
\author[$2$]{Francisco Perdigon Romero}
\author[$1$]{Eric Granger}

\affil[$1$]{Laboratoire d’imagerie, de vision et d’intelligence artificielle (LIVIA)\\ Dept. of Systems Engineering, École de technologie supérieure, Montreal Canada}
\affil[$2$]{GAIA Montreal, Ericsson Canada}

\maketitle


\begin{abstract}
Semi- and weakly-supervised learning have recently attracted considerable attention in the object detection literature since they can alleviate the cost of annotation needed to successfully train deep learning models. State-of-art approaches for semi-supervised learning rely on student-teacher models trained using a multi-stage process, and considerable data augmentation. Custom networks have been developed for the weakly-supervised setting, making it difficult to adapt to different detectors. 
In this paper, a weakly semi-supervised training method is introduced that reduces these training challenges, yet achieves state-of-the-art performance by leveraging only a small fraction of fully-labeled images with information in weakly-labeled images. In particular, our generic sampling-based learning strategy produces pseudo ground-truth (GT) bounding box annotations in an online fashion, eliminating the need for multi-stage training, and student-teacher network configurations. These pseudo GT boxes are sampled from weakly-labeled images based on the categorical score of object proposals accumulated via a score propagation process. 
Empirical results\footnote{Our code is available at: \href{url}{https://github.com/akhilpm/SemiWSOD}} on the Pascal VOC dataset, indicates that the proposed approach improves performance by 5.0\% when using VOC 2007 as fully-labeled, and VOC 2012 as weak-labeled data. 
Also, with 5-10\% fully annotated images,  we observed an improvement of more than 10\% in mAP, showing that a modest investment in image-level annotation, can substantially improve detection performance.
\end{abstract}

\section{Introduction}
\label{sec:introduction}
Recent advances in object detection are mainly due to deep learning models trained using a massive amount of densely annotated images\cite{fast_rcnn-Girshick-2015, detr-Carion-2020, fpn-Lin-2017, yolo_9000-Redmon-2017, rfcn-Dai-2016, fcos-Tian-2019, ssd-Liu-2016}. The main development bottleneck of such detector models is the requirement for dense annotation of  each image, i.e., the bounding box and class label of each instance present in an image. Therefore, producing a curated and well balanced dataset with annotations is a costly undertaking, often requiring expert knowledge. These challenges have motivated research on learning paradigms to train object detectors with reduced annotation. Popular approaches in this direction can be broadly categorized into semi-supervised learning\cite{humble-teacher-Yang-2021, soft-teacher-Xu-2021}, weakly supervised learning\cite{oicr-Tang-2017, instance-aware-Ren-2020}, active learning\cite{deep-active-roy-2018, deep-active-li-2021}, and few shot learning\cite{defrcn-Qiao-2021}. In semi-supervised settings, some approaches additionally used weak image-level labels\cite{wssod-Fang-2021} and point annotations\cite{points_weakly_semi_sup-chen-2020}.

Common steps in semi- and weakly-supervised learning consist of first extracting pseudo GT boxes, and then self-training the detector with these pseudo annotations\cite{oicr-Tang-2017, points_weakly_semi_sup-chen-2020, humble-teacher-Yang-2021, instance-aware-Ren-2020}. However, this often results in multiple stages of training, leading to learning difficulties that vary according to the settings. In addition to that, the extraction of pseudo GT boxes often involves a precise thresholding of detection scores in order to find a good trade-off between precision and recall, which 
can lead to unstable training\cite{csd-Jeong-2019}. In semi-supervised learning, the self-training targets are obtained from a first stage of training with only the fully annotated samples, which are less in number so prone to overfitting\cite{points_weakly_semi_sup-chen-2020}. In weakly-supervised learning, a custom MIL (Multiple Instance Learning) based detector must first be trained to generate these pseudo annotations, and then a standard fully-supervised detector is self-trained with the pseudo targets\cite{oicr-Tang-2017, instance-aware-Ren-2020}. Therefore, multiple training stages are required, and 
multiple detection architectures need to be trained. Also the pseudo target extraction must be performed offline, resulting in a training process that is not end-to-end.

In this paper, we propose an online self-training method for weakly semi-supervised object detection that exploits a small fraction of fully-annotated images, along with the remaining weakly-annotated ones, for efficient training. 
In this settings, we propose a sampling strategy where pseudo GT boxes are sampled using the semantics accumulated on object proposal regions. For images with real bounding box annotation, we compute the normal classification and localization loss to update the network. For images with weak annotations, we sample a set of pseudo GT boxes for each given category present in the image, and classify and regress those, allowing to use a standard detector again. The pseudo GT boxes are sampled from object proposals extracted from weakly-labeled images. Thus, during training, our network learns fully- and weakly-annotated images in a single stage, utilizing any backbone object detection architecture. Unlike other self-training based model for semi-supervised object detection, our method does not require thresholding the detection scores for defining the number of pseudo GT boxes, which translates to less effort in the training process. For sampling, we make use of the category specific score of the object proposals. Category specific score of object proposals are accumulated by propagating the scores of detection boxes to object proposals based on their Intersection over Union (IoU). For experimental validation, the proposed approach is compared against state-of-art methods for semi-supervised learning on the Pascal VOC dataset. 

\section{Related work}
\label{sec:related-work}
\subsection{Supervised Object Detection:}
Models for object detection can be broadly classified into two-stage\cite{faster_rcnn-Ren-2015, fpn-Lin-2017} and one-stage \cite{yolo_9000-Redmon-2017, ssd-Liu-2016} object detectors. In both cases, supervised object detection requires class and bounding box labels for each instance present in the image\cite{rcnn-Girshick-2016}.
Two-stage object detectors have a first stage that extracts RoIs (candidate object regions) whose reliability as a potential candidate region is quantified by their objectness score. Earlier approaches extract these RoIs using low-level image features in R-CNN\cite{rcnn-Girshick-2016}, Fast R-CNN\cite{fast_rcnn-Girshick-2015}, etc. Later, end-to-end two-stage models emerged as more accurate detectors, where an additional learnable head called RPN (Region Proposal Network) is used to regress candidate regions\cite{faster_rcnn-Ren-2015, fpn-Lin-2017, rfcn-Dai-2016}.  
In contrast, one-stage object detectors avoid the RoI extraction stage, and classify and regress directly from the anchor boxes. They are  generally fast and applicable to real-time object detection\cite{yolo_9000-Redmon-2017, yolo-Redmon-2016, ssd-Liu-2016}. Though the two-stage detectors are typically slower compared to their one-stage counterparts, extracting reliable candidate region in the first stage provides an edge in terms of localization accuracy. Recently, anchor-free detectors\cite{fcos-Tian-2019, centernet-Duan-2019} have grown in popularity since they dissociate from hand-designed anchor box selection and matching process, and provide even faster detectors.

\subsection{Semi/Weakly-Supervised Object Detection:} 
Semi-supervised object detectors rely on a small subset of labeled images and a large collection of unlabeled images to train a detector\cite{humble-teacher-Yang-2021}. They have been explored in many different forms. Recently, detectors trained in a student-teacher fashion have gained popularity in the research community\cite{soft-teacher-Xu-2021, stac-sohn-2015}. In this setting, a teacher model is trained first using the available annotated data, and then used to produce pseudo-labels for unlabeled data. Once the pseudo-labels are obtained, a student model is trained using the combined dataset. Though our proposed method also make use of the concepts of pseudo-labeling, our approach differs from these methods because we don't have two separate networks, and multiple training stages, making it a simpler solution for semi-supervised detection.

Weakly-supervised detectors, on the other hand, only rely on image-class labels\cite{oicr-Tang-2017, wsddn-Bilen-2016, wccn-Diba-2017}. Research in weakly-supervised methods have progressed greatly in recent years\cite{instance-aware-Ren-2020}. However, a fundamental problem is their inability to distinguish the entire object from its parts and context. As a result, the detectors often produce bounding boxes on discriminative object parts, failing to distinguish the object boundaries when multiples objects of the same class are spatially adjacent. They produce imprecise localization that do not differentiate object boundaries from its context. Our proposed method also learns from a vast majority of weakly-labeled images, but, combined with a few fully-annotated images, providing better objectness distillation from the fully- to weakly-annotated images. We argue that this is the most practical setting, as we can typically annotate a small number of images with bounding boxes, and supply the remaining with weak annotations, or no annotations at all.
Some approaches also attempt to leverage the additional, easy to obtain weak supervision for the unlabeled data. In particular, they used weak image-level labels\cite{wssod-Fang-2021} or point annotations\cite{points_weakly_semi_sup-chen-2020} for the unlabeled images. Though this required additional annotation effort, they are easy to obtain, and may provide substantial improvements. Our approach also relies on weak image-class labels.

\subsection{Self-Training:}
Self-training has emerged as a popular approach in weakly and semi-supervised object detection\cite{oicr-Tang-2017, instance-aware-Ren-2020, points_weakly_semi_sup-chen-2020}. In these approaches, a detector is first trained on the available data and then the obtained detections from the first training are used for training a refined model. This procedure can be repeated often multiple times. 
However, these methods perform self-training with pseudo labels obtained offline, e.g., a Faster R-CNN network training with detection result of weakly-supervised detectors as pseudo labels\cite{oicr-Tang-2017, instance-aware-Ren-2020}. In contrast, our method performs self-training using the pseudo labels for weakly labeled images, but in an online fashion. In particular, for each iteration, our approach samples pseudo GT boxes for each given weak label using the accumulated semantics from a score propagation block. Given this online nature of the self-training, the training is a simple one-stage process.
\section{Proposed Method}
\label{sec:proposal}

\begin{figure*}[h!]
  \centering
  \vspace{-0.5cm}
  \includegraphics[ width=0.9\linewidth]{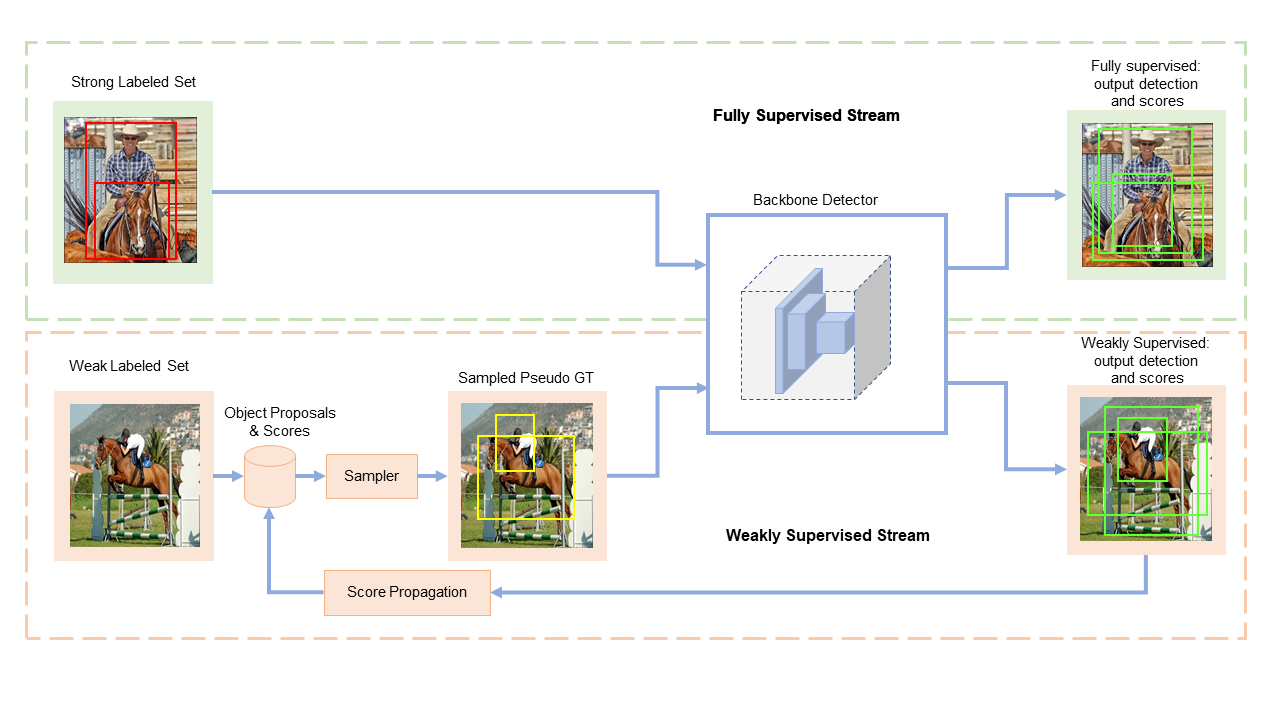} \vspace{-0.5cm}
  \caption{\textbf{Proposed Method for Weakly Semi-Supervised Training.} For fully-labeled  images, our approach uses the backbone detector without any modifications. For weakly-labeled images, our approach samples pseudo GT boxes from a set of proposals. The probability of sampling a proposal is proportional to the score given by the detector to the most overlapping detection. In this way, the sampling approach focuses more and more to the objects of interest in the image.}
  \label{fig:overall_system}
\end{figure*}

In this section, we introduce a learning technique for semi-weakly supervised object detection. Fig \ref{fig:overall_system} illustrates the overall system design. The backbone of our system can be any fully supervised architecture -- we used Faster-RCNN\cite{faster_rcnn-Ren-2015} in our experiments. For every input image, the detector is employed in a different way, depending on the available annotation level. 
For the fully-labeled images, we perform a normal forward-backward cycle by taking the real GT annotations provided. For the weakly-labeled images, we use an importance sampling approach to select the most likely bonding boxes for each class in the image, to use them for training as a pseudo ground-truth. 
This allows us to use the same detector employed for the fully-labeled images also for the weakly-labeled ones. 

When combining weakly and fully supervised learning, we need to determine the right importance to associate to the two learning tasks.
For doing that, we rely on a hyper-parameter that defines the sampling ratio between the fully- and weakly-labeled images. This allows to associate a different level of importance to both tasks without changing the actual losses defined in the detector code. The rest of this section introduces the learning steps of both strongly and weakly-annotated category. Then, we present how their sampling ratio is applied within the training process. 

\subsection{Learning with Strong Annotations:}
For the images that are strongly annotated (.i.e bounding box annotation for each object present), the learning step is straightforward. 
Given an input image $I$, the ground truth (GT) annotations are defined with the bounding box positions $\mathcal{B}=\{b_0,b_1,\cdots\ b_N\}$ and corresponding classes $\mathcal{C}=\{c_0,c_1,\cdots\ c_N\}$. $b =(x_0,y_0,x_1,y_1)  $ is a vector with 4 values that represents for instance the top left and bottom right corner of a box, while $c \in \mathcal{C}$ is a discrete value that represents the object category.
In our experiments we use faster RCNN as detector \cite{faster_rcnn-Ren-2015}, and thus use a loss as:
\begin{equation}
\begin{aligned}
    L_{F} = \sum_{j \in \mathcal{D}} \sum_{i \in \mathcal{B}} \frac{1}{N_{cls}}   L_{cls}(f_{c_i}(d_j,I)) \\ + \lambda \frac {1}{N_{reg}}  L_{reg}(c_i,d_j, b_i) ,
    \label{equ:softmax}
\end{aligned}
\end{equation}
For each GT bounding box $b_i$ it generates a loss based on the scores $f_{c_i}$ and overlap of the obtained detections $d_j$.
$L_{cls}$ and $L_{reg}$ denote the classification and localization loss, respectively. 
$N_{cls}$ and $N_{reg}$ are the normalization factors which depends on the number of foreground and background RoIs considered. $\lambda$ is a hyperparameter that controls the relative importance of the classification and localization loss. 
Note that the exact form of loss can vary according to the fully supervised detector architecture used in the model, but our approach is independent of the specific fully supervised loss.

\subsection{Learning with Weak Annotations:}

In case of weak supervision, we know the object classes that are present in the image $c$, but not the bounding box locations $b_i$. Thus, the general approach of weakly supervised models is to learn during training what are the regions of the image that are more likely to contain the object of interest.
A typical approach for doing that is to compute the classification
loss for a given class $c$ and an image $I$ as a weighted sum of scores:
\begin{equation}
    L_{cls} \Big( \underbrace{\sum_l w_l f_{c}(p_l,I)}_{h} \Big),
    \label{equ:weakcls}
\end{equation}
where $f_{c}(p_l,I)$ is the score of the detector for ground truth class $c$ and object proposal $p_l \in \mathcal{P}$ of to the image $I$, and $w_l$ is a normalized weight associated with each object proposal $p_l$.
The weights associated with each proposal $w_l$ for a image $I$ are computed as normalized scores:
\begin{equation}
    w_l = \frac{\exp \{ \frac{f_{c^*}(p_l,I)}{T} \} }
            {\sum_j \exp \{ \frac{f_{c^*}(p_j,I)}{T} \} },
    \label{equ:weakcls_weights}
\end{equation}
with $T$ being the temperature parameter that defines the sharpness of the weight distribution and is a hyper-parameter of the learning approach.
In this way, boxes with higher score will have more impact on the learning and the learning will focus more and more on the locations of the image that are more likely to contain the object of interest.
While this formulation works well, it is computationally expensive because it has to evaluate at each training iteration and for each image all box locations $p_l$.

Here, we propose to approximate the weighted sum of scores $h$ with Monte Carlo sampling, in which instead of computing the sum over entire bounding box locations $l$, we uniformly sample $K$ boxes:
\begin{equation}
    h \approx \hat{h} = \sum_{k \sim \mathcal{U}} \theta_k f_{c}(p_k,I), 
\end{equation}
where $\theta_k$ is the weight associated with the bounding box $k$.
This allows us to compute only $K$ evaluations of the expensive $f$, while using an unbiased estimation of the weakly supervised scoring function. We call these samples pseudo GT bounding boxes because as we will see in the next section, these samples can be pass to the detection algorithm as ground truth annotations. This allows our algorithm to use any off-the-shelf object detector without modification.

However, when sampling, the weights $\theta_k$ associated with each bounding box score cannot be computed directly because it is the normalized version of the object score $f_{c}(p_k,I)$ for the box $p_k$, but we do not have all scores for the normalization factor, as we sampled only a few boxes. Instead, for each image $I$ we keep in memory the scores $s_l$ associated with the bounding box $p_l$ and update only the score of the $K$ sampled bounding boxes.  Then, $\theta_k$ will be computed as the normalized version of $s_k$:
\begin{equation}
    \theta_k = \frac{ \exp\{ \frac{s_k}{T} \} }{\sum_j \exp \{ \frac{s_j}{T} \} }
    \label{equ:theta}
\end{equation}
As $\theta_k$ approximates $w_l$,
we see that when the scores $f$ do not change anymore (i.e. at convergence), $\hat{h}$ becomes $h$. Therefore $\hat{h}$ is an unbiased estimation of $h$.
While this approach would work, sampling uniformly any possible image bounding box proposal $p_l$ would make the learning very slow because most of the time the sample would not come from the object of interest.
Instead, in this work we use an importance sampling approach. We use $\theta_k$ as sampling probabilities associated with a multinomial distribution $\mathcal{M}(\theta_k)$ to sample bounding boxes, so that the bounding box proposal $p_k$ would have a probability $\theta_k$ to be sampled.
In this case, in order to maintain the same estimation of $h$ we need to divide by the sampling probability $\theta_k$. Thus the final estimation of $h$ will be:
\begin{equation}
    \hat{h} = \sum_{k \sim \mathcal{M}(\theta_k)} f_{c}(p_k,I),
\end{equation}
which is again an unbiased estimator of the classification score of an image, but with lower variance. 
Thus, the final weakly supervised loss $L_W$ for an image $I$ is the same as in Eqn.(\ref{equ:softmax}), but with the ground truth bounding boxes $b_i^*$ changed to the pseudo GT boxes sampled from the object proposals $\mathcal{P}$.

\subsection{Combining Strong and Weak Annotation:}
While the fully supervised object detector uses ground truth boxes that are correct, the weakly supervised counterpart estimates the object box location during training and therefore the estimation can be noisy.
Thus, when learning with strong and weak labels we might want to set a hyper-parameter value that balances the relative importance of the two losses. In this case the final loss is $ L = L_F + \lambda L_W $.
As the aim of this approach is to avoid to directly modify the loss on the detector, in this case we express the weight $\lambda$ with a sampling ratio.  
Thus, we use a ratio parameter $r$ that controls the amount of training data from the fully and weakly labeled pool of data. For instance $r=0.6$ means that we sample with probability $0.6$ from the pool of the fully-labeled samples and $0.4$ from the weakly-labeled samples. 
This approach has the advantage of not changing the internals of the underlying detector such as loss function or additional regularization etc. With this design, we can feed both the fully annotated and weakly annotated images in parallel to the model and train it in a single stage. 
\subsection{Learning Algorithm}
In this section we present in more detail the different parts of the proposed learning algorithm as summarized in Alg.~\ref{alg:1}. For the sake of simplicity the algorithm is shown for the case of a single image $I$, but it could be trivially extended to a batch of images.

For supervised samples our algorithm uses directly the bounding box annotations $\mathcal{B}$ and the corresponding classes $\mathcal{C}$ for inference (\verb|detect|).
For weak supervision, the inference is performed on pseudo GT annotations ($\hat{\mathcal{B}},\hat{\mathcal{C}}$) that are obtained by sampling object proposals (\verb|sample|). Then, the obtained detections $\mathcal{D}$ and scores $\mathcal{S}^D$ are used to \verb|update| the proposal scores ($\mathcal{S}^P$).
In both cases, the obtained detections $\mathcal{D}$ and scores $\mathcal{S}^D$ are used to compute the loss $L$ and update the recognition model (\verb|backprop|). 

\begin{algorithm}
\caption{Semi-Weakly supervised learning with Pseudo GT}
\KwIn {Image: I, GT:$(\mathcal{B},\mathcal{C})$ proposals and scores: $(\mathcal{P},\mathcal{S}^P)$}
  \uIf{$\mathcal{B} \neq \emptyset$}{
    \emph{fully supervised} \;
    $\mathcal{D},\mathcal{S}^D$ = {\texttt{detect}}(I, $\mathcal{B},\mathcal{C}$)\;
  }
  \Else{
   \emph{weakly supervised}\;
   $\hat{\mathcal{B}},\hat{\mathcal{C}}$ = {\texttt{sample}} ($\mathcal{P},\mathcal{S}^P,\mathcal{C}$) \;
   $\mathcal{D},\mathcal{S}^D$ = {\texttt{detect} (I, $\hat{\mathcal{B}},\hat{\mathcal{C}}$) \;
   $\mathcal{S}^P$ = {\texttt{update}} ($\mathcal{P},\mathcal{D},\mathcal{S}^D$)\;
  }}
  {\texttt{backprop}( I , $\mathcal{B},\mathcal{C},\mathcal{D},\mathcal{S}^D$)}
\label{alg:1}
\end{algorithm}

\subsubsection{Detection inference}(\verb|detect|)
This can be any detection algorithm that takes as input the ground truth annotations ($\mathcal{B,C}$) for an image $I$ and returns detections $\mathcal{D}$ with associated scores $\mathcal{S}^D$ for all classes. Notice that, in general, ground truth annotations $\mathcal{B}$ are not needed to perform inference. However, in many detectors, these are used to limit the detections computation nearby the GT annotations.

\subsubsection{Sampling Pseudo GT} 
\begin{figure}
\centering
\begin{tabular}{c @{\hspace{0.5ex}} c @{\hspace{0.5ex}} c @{\hspace{0.5ex}} c}
\includegraphics[width=0.23\linewidth]{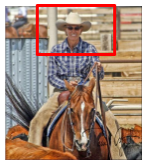} &
\includegraphics[width=0.23\linewidth]{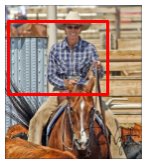} &
\includegraphics[width=0.23\linewidth]{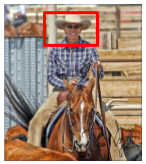} &
\includegraphics[width=0.23\linewidth]{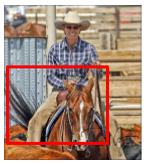} \\
 Epoch 1 & Epoch 5 & Epoch 10 & Epoch 15\\
 \includegraphics[width=0.23\linewidth]{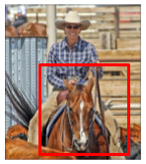} &
\includegraphics[width=0.23\linewidth]{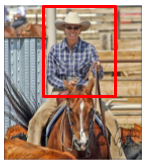} &
\includegraphics[width=0.23\linewidth]{images/sampling_evolution/sampling_epoch_20.png} &
\includegraphics[width=0.23\linewidth]{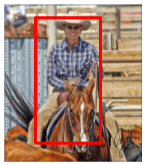} \\
 Epoch 20 & Epoch 25 & Epoch 30 & Epoch 35\\
 \end{tabular}
 \caption{Evolution of the Pseudo GT sampling. While in the first iterations of the training, bounding boxes are samples almost randomly (exploration), after some training, the algorithm learns to sample only from meaningful locations (exploitation). }
 \label{fig:sampling_evolution}
\end{figure}

(\verb|sample|)
For each proposal $p_l \in \mathcal{P}$ we have the corresponding classification score $s_{l,c}$ for a given class $c$. This score is accumulated based on the detector output via the score propagation step which will be explained in the next section. 
For each class present in an image, we consider the scores of all proposals $\mathcal{P}$, denoted by $\mathcal{S}^P$ and sample K boxes based on the multinomial distribution $\mathcal{M}$ with probabilities $\theta$ computed as in Eqn.\ref{equ:theta}. 
Figure \ref{fig:sampling_evolution} illustrates an example sampling process during the training phase for the person category. It can be observed that, though in the beginning we sample pseudo GT boxes randomly from the image, it converges to meaningful locations for the person category in the later stages.

\subsubsection{Score Propagation}(\verb|update|)
Score propagation is the component which updates the score values $\mathcal{S}^P$ of the object proposals $\mathcal{P}$.
If the output bounding boxes of the detector $\mathcal{D}$ would correspond to the object proposals $\mathcal{P}$ 
, we could directly copy the detection values to our pool of proposals.
Instead, as in modern detectors the output detections $\mathcal{D}$ are generated by a regression, here we propose a method to propagate the scores from the output detection $\mathcal{D}$ to the scores $\mathcal{S}^P$ of our object proposals that are then sampled as pseudo GT.  
During learning, the proposals will accumulate scores from their overlapping boxes produced by the detector. In our design, we define the score propagation according the percentage of overlap between a proposal $p_l$ and a detection $d_l$. This will help the proposals to aggregate the detection scores of its neighborhood region during learning.

Initial score values are initialized to 0. Then, during learning, their scores will be updated based on detection scores.
We explored several criteria for propagating the score and observed that propagating scores from the maximum overlapping detection boxes helps the model to collect better semantics for the region.
Thus, we define $\gamma$ as the maximum intersection over union between proposal $p$ and all detection boxes $d \in \mathcal{D}$:
$\gamma =\max_{d \in \mathcal{D}} \frac{p \cap d}{p \cup d}$.
So, for each proposal $p_l$ present in the image, we propagate its score $s_{l,c}$ proportional to $\gamma$:
\begin{equation}
\label{score_prop_formula}
s_{l,c} = (1 - \gamma) s_{l,c} + \gamma \cdot s_{d,c},
\end{equation}
where $s_{d,c}$ is the score of the maximum overlapping detection box $d$ for category $c$. In this way, scores associated to the proposal $l$ with high overlap with the detection $d$ will receive a strong update, while scores of proposals with low overlap will not influence the stored score $s_{l,c}$.


\subsubsection{Updating the model}(\verb|backprop|)
The update of the detector model is performed as usual. The loss $L_{F}$ that evaluates the difference between the ground truth annotations (or pseudo GT) $(\mathcal{B},\mathcal{C})$ and the obtained detections ($\mathcal{D},\mathcal{S}^D$) is computed. Then, the model parameters are updated with standard back-propagation. 

\section{Extensions}

\subsection{Reducing Proposals with Class Activation Maps:}
In our basic model we use around 2000 object proposals obtained from the selective search algorithm \cite{selective_search-Sande-2011} in order to have a high detection recall. However, keeping so many proposals, means that we need to keep a large set of scores. This will make the algorithm slow and more noisy in the beginning of the training as there are many possible regions to explore. 
In the experiments, we tested to use a Class Activation Map (CAM) model \cite{gradcam-Selvaraju-2017} to reduce the initial number of proposals.
In practice, for each image and for each class present in an image, we extract its CAM. Then, for each CAM region, only the proposals that overlap at least $\rho$ with that CAM region are kept. The final set of proposals will be the union of the proposals selected for each class.

\subsection{From Weakly to Fully Semi-Supervised Learning:}
We extend the weakly semi-supervised approach proposed to a normal semi-supervised approach, in which part of the data is fully supervised and the rest has not supervision at all. For doing that, we keep the same structure of the previous approach, but for the non-annotated images, instead of using ground truth labels, we use the labels provided by the image classifier already used to compute the CAMs. As we will see in the experiments, this will provide almost the same level of performance as the semi-weakly supervised but without the need of weak labels.  


\section{Experiments}
\label{sec:experiments}

\subsection{Experimental Setup:}

\noindent \textbf{Datasets and evaluation metric:} The effectiveness of our proposed method is evaluated on 
VOC07 and VOC12~\cite{voc-Everingham-2010}. 
We perform ablation experiments on VOC07. In order to test the performance of our algorithm with different amount of fully annotated and weakly annotated training data, we use the trainval split of VOC07 into different percentages of fully annotated and weakly annotated set. We used 0\%, 5\%, 10\%, and 20\% of the images with bounding box annotations in this study and the rest with image-level labels. The images are sampled randomly to create the fully annotated and weakly annotated split. For all experiments with VOC dataset, we used the test set for evaluation. The standard VOC AP metric(AP 50) is used to measure the performance of the model. 
Finally to compare with other weakly and semi-supervised approaches, we trained using VOC 2007 trainval as the fully labeled set, and VOC 2012 trainval as the weakly and unlabeled sets. 

\noindent \textbf{Implementation details} In most of our experiments, we used VGG16\cite{vgg16-simonyan-2015} as the CNN backbone, pre-trained on the ImageNet\cite{imagenet-Russakovsky-2015} dataset. We also reported results with a ResNet101 backbone\cite{resnet-He-2016} in order to present of fair comparisons with other state-of-the-art methods. The backbone detector used in our study is Faster R-CNN\cite{faster_rcnn-Ren-2015}. The whole network is trained end-to-end using stochastic gradient descent(SGD) with a momentum of $0.9$ and a weight decay of $0.0005$. The initial learning rate is set to 1e-2 and decayed at epochs [5,10] by a factor of 10. We trained the model for $20$ epochs with a batch size of 8. The temperature parameter $T$ for the multinomial distribution used for sampling is set to 2.5. From an image for each class present, we sample $K=5$ object proposals as pseudo GT during training. 

During training, the shorter edge of the input images are randomly rescaled within \{480,576,688,864,1200\}. We only use horizontal flipping for data augmentation. Object proposals are extracted using the selective search algorithm \cite{selective_search-Sande-2011}. Typically, from an image, up to 2000 object proposals are extracted for good recall of all the object instances. 
Images are normalized with $\textrm{mean} = [0.485, 0.456, 0.406]$ and $\textrm{std} = [0.229, 0.224, 0.225]$, as in ImageNet training \cite{imagenet-Russakovsky-2015}. The network is trained on NVIDIA V100 GPU with 32GB memory.

\subsection{Ablation Study:}
Several ablation studies are conducted in order to asses the individual components of our proposed model. First, we study the impact on performance of different ways of defining the sampler and score propagation module. Then, we analyze the impact of learning with a growing amount of annotations. Finally, we study the contribution of different types of errors made by the model. All of the these studies are conducted on PASCAL VOC 2007 by training the model using its trainval set, and testing on its test set. We used 10\% bounding annotations in this analysis, while the rest of the images are weakly annotated.

\subsubsection{Sampler and score propagation}
\begin{figure}
\centering
\begin{tabular}{l @{\hspace{-8ex}} c}
\begin{subfigure}{.95\linewidth}
\centering
\includegraphics[width=.2\linewidth]{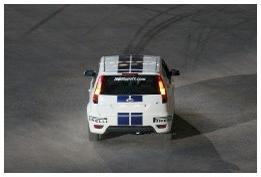}\vspace{0.15\baselineskip}
\includegraphics[width=.2\linewidth]{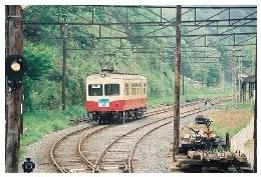}
\includegraphics[width=.2\linewidth]{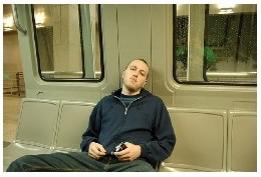}
\includegraphics[width=.2\linewidth]{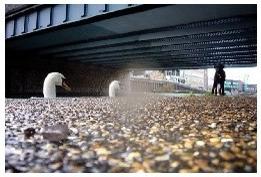} 
\includegraphics[width=.2\linewidth]{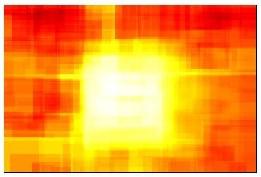}\vspace{0.15\baselineskip} 
\includegraphics[width=.2\linewidth]{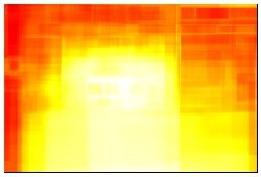}
\includegraphics[width=.2\linewidth]{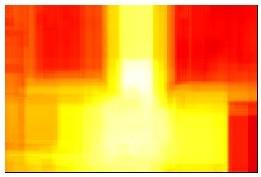}
\includegraphics[width=.2\linewidth]{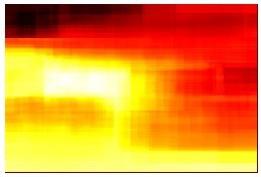}
\includegraphics[width=.2\linewidth]{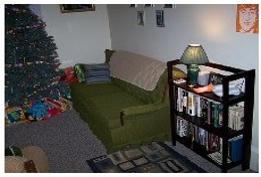}\vspace{0.15\baselineskip} 
\includegraphics[width=.2\linewidth]{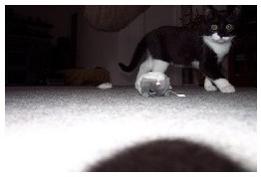}
\includegraphics[width=.2\linewidth]{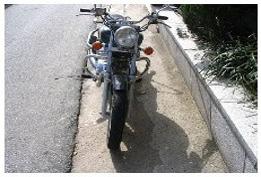}
\includegraphics[width=.2\linewidth]{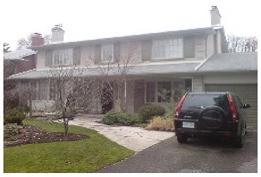}
\includegraphics[width=.2\linewidth]{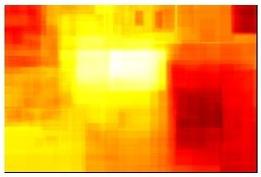}\vspace{0.15\baselineskip}
\includegraphics[width=.2\linewidth]{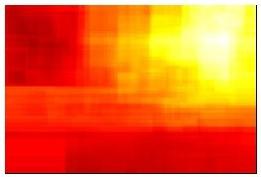} 
\includegraphics[width=.2\linewidth]{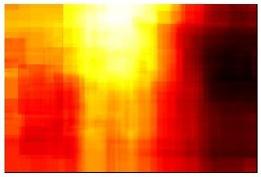}
\includegraphics[width=.2\linewidth]{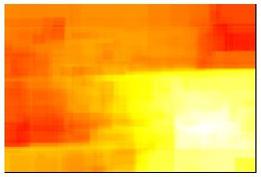}
\end{subfigure} &

\begin{subfigure}{.15\textwidth}
\centering
\includegraphics[ width=.3\textwidth]{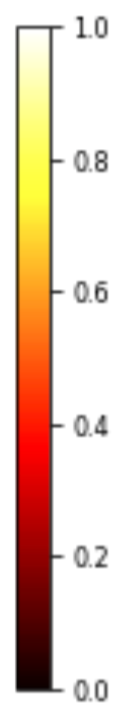}
\end{subfigure}
\end{tabular}
\caption{Examples of heatmap of sampler scores for images belonging to different Pascal VOC categories.}
\label{fig:heatmap}
\end{figure}

To understand whether the sampler is learning a meaningful object location, we analyze the heatmap produced by the score distributions of the object proposals for a given class. To obtain the heatmap, for each pixel location, the scores from all object proposals covering that pixel are added, and then normalized by the number of object proposals covering that pixel. Fig.~\ref{fig:heatmap} shows some examples of heatmaps. It can be observed that active regions of heatmaps correlate well with object locations, and hence the sampler is finding meaningful semantic information through sampling and score propagation. We also notice that for small objects (ducks on the top right image) or objects with a recurrent background (train), the sampler selects not only the object of interest but also some background. However, this is a common problem of all weakly supervised approaches.

Score propagation can be designed in many ways based on, e.g., all detection boxes, or a selected set boxes matching some quality criteria. In this study, 3 settings are considered: (1) score propagation from all detection boxes, (2) from the maximum overlapping boxes, and (3) from the maximum overlapping boxes when the overlap is above a threshold $t$. We found that $t=0.3$ provides the best performance. Table \ref{table:score_propagation} summarizes the results from this study on VOC 2007 dataset. The model is trained using different 10\% split on its trainval set, and evaluated on the test set. It can be observed that propagating scores from the maximum overlapping detection box of each proposal provides the highest mAP accuracy. When the overlap is above a threshold $t$ imposes more quality constraints for score propagation, and improves the results. Score propagation from all detection boxes does not perform well, although it can provide a smoother update to the object proposal scores. This may be due to the distribution of the high scores over a large area when all detection boxes are propagating their scores. This results in incorrect sampling of over-sized proposals, especially for smaller objects.

\begin{table}[!t]
    \caption{Analysis of score propagation strategies. The performance is measured on a weakly semi-supervised model using with 10\% full annotations and remaining weakly-labeled images on the VOC 2007 dataset.}
    \centering
    \begin{tabular}{ l  c }
        \hline
        \textbf{Score Propagation Strategy} & mAP  \\
        \hline
        \hline
        Propagate from all boxes & 57.2 \\
        \hline
        \shortstack{Propagate from max-overlapping boxes} & 58.3 \\
        \hline
        \shortstack[l]{Propagate from max-overlapping \\ boxes when IOU $> \theta$} & 60.3 \\
        \hline
    \end{tabular}
    \label{table:score_propagation}
\end{table}

\subsubsection{Impact of fully annotated images}

\begin{figure}[!b]
  \centering
  \includegraphics[height=5cm, width=0.95\linewidth]{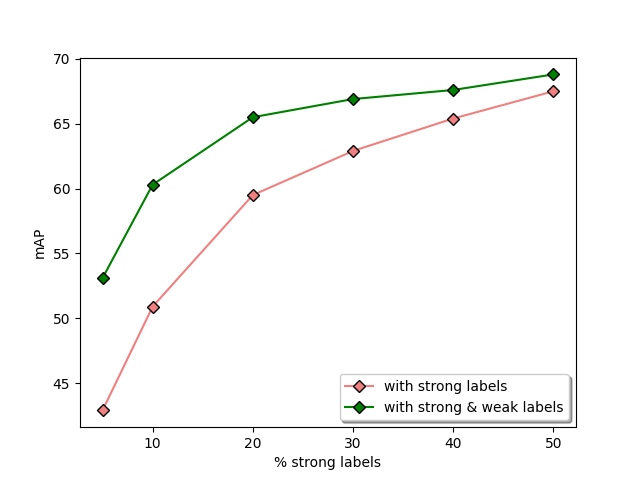} 
  \caption{The impact on mAP performance of adding more fully annotated images during training on the VOC 2007 dataset.}
  \label{fig:label_vs_mAP}
\end{figure}
In this experiment we compare the performance of a detector baseline trained only with strong labels (red line) with a model trained with strong and weak labels using our sampling approach (green line). Figure \ref{fig:label_vs_mAP} shows the results from this study.
As expected, the gain of our model is more significant when the amount of strong labels is reduced.
For instance, with $5\%$ of strong labels, our model improves over the baseline by $10$ points. When increasing the percentage of strong labels, the gain reduces until a few points when using $50\%$ of strong labels.
This experience shows how our approach is particularly useful when using a very reduced amount of fully labeled data and the rest is weakly labeled. In this setting, our model can approach the performance of a fully supervised model, but with much fewer annotations.



\subsubsection{Impact of the ratio parameter}
We analyze the importance of the ratio parameter $r$ for balancing the number of fully and weakly annotated training images. In Table \ref{table:ratio_impact}, it can be observed that without this balancing, the detection performance is even worse than the settings where only annotated images are used. With the proper ratio balancing ($r=0.7$), the mAP performance of the detector significantly outperforms the baseline using only fully supervised images. Thus 
by tuning this ratio parameter, we can  effectively leverage the large pool of weakly-annotated images. One of the appealing property of this strategy is that it does not require any change to the model architecture or loss function.
\begin{table}[!t]
    \caption{Impact of the ratio of fully to weakly annotated images.}
    \centering
    \begin{tabular}{ l  c }
        \hline
        \textbf{Settings} & mAP  \\
        \hline
        \hline
        \shortstack{$10\%$ fully annotated images} & 50.9 \\
        \hline
        \shortstack[l]{10\% fully annotated and remaining weakly \\ annotated(without ratio balancing)} & 47.5 \\
        \hline
        \shortstack[l]{10\% fully annotated and remaining weakly \\ annotated(with ratio balancing)} & 60.3 \\
        \hline
    \end{tabular}
    \label{table:ratio_impact}
\end{table}

\subsubsection{Impact of CAM proposals} 
Table~\ref{table:cam_proposal_impact} shows the impact on performance when considering the overlap between object proposals and CAM of the relevant class  during sampling. The CAM is obtained by training a vgg16\cite{vgg16-simonyan-2015} network on the multi-label VOC 2007 image-level labels. Then the overlap of selective search proposals\cite{selective_search-Sande-2011} to the CAM of all classes present in the image is computed. Based on the overlap, the object proposals without sufficient overlap to the CAM, which are perhaps from the image background region, are ignored. This results in a slight loss of recall, but an improvement in terms of the mAP, especially with few fully annotated images, due to reduction of noisy proposal regions that could  misguide the sampler. In practice, we used an overlap threshold of 0.1 which result in a 5\% reduction of recall, but the average number of proposals is reduced 4 times to approximately 500 object proposals per image. From table \ref{table:cam_proposal_impact}, it is clear that filtering noisy proposals using CAM brings improvement in mAP. But the impact of the CAM proposals reduces with the availability of more fully annotated images. This is in accordance to the general facts that with more annotations, the appearance model will be more accurate and hence, the model itself will be powerful enough to distinguish the object boundaries very well.

\begin{table}[!b]
    \caption{Impact on performance when using proposals filtered by a CAM model. }
    \centering
    \begin{tabular}{ c  c  c }
        \hline
        \textbf{\shortstack[l]{\% images with \\ bounding box \\ annotation}} & \textbf{\shortstack[l]{mAP without \\ CAM proposals}} & \textbf{\shortstack[l]{mAP with \\ CAM proposals}} \\
        \hline
        \hline
        0\% & 27.2 & 35.5 \\
        5\% & 48.4 & 53.1 \\
        10\% & 57.6 & 60.3 \\
        20\% & 64.6 & 65.5 \\
        \hline
    \end{tabular}
    \label{table:cam_proposal_impact}
\end{table}

\subsubsection{Loss in performance}
\begin{figure}[htp]
  \centering
  \includegraphics[height=6.5cm, width=0.6\linewidth]{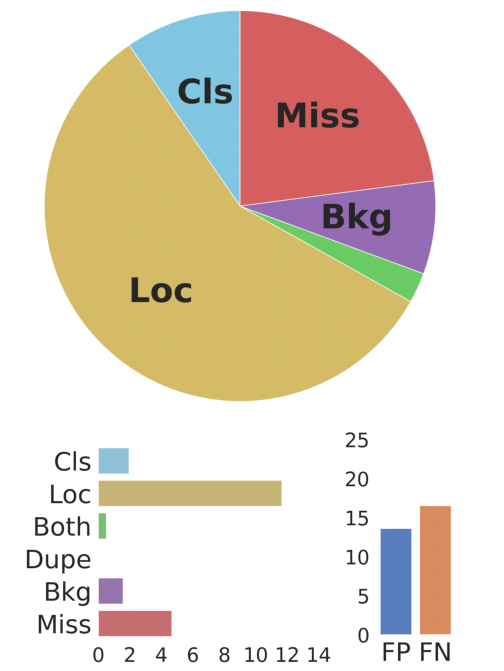} 
  \caption{\textbf{Evaluation of performance loss.} TIDE\cite{tide-Bolya-eccv2020} Evaluation of detection results. Error types are: \textbf{Cls}: localized correctly but classified incorrectly, \textbf{Loc}: classified correctly but localized incorrectly, \textbf{Both}: both cls and loc error, \textbf{Dupe}: duplicate detection error, \textbf{Bkg}: detected background as foreground, \textbf{Miss}: missed ground truth error.
}
  \label{fig:tide_evaluation_voc2007}
\end{figure}

We also analyze the distribution of the error of our model using the TIDE\cite{tide-Bolya-eccv2020} evaluation tool (see Figure \ref{fig:tide_evaluation_voc2007}). It can be observed that the localization error contributes the most towards the overall errors made by our detection model. This is expected, since there is a large fraction of the images without bounding box labels, so the objectness distilled from a small fraction of fully annotated images is insufficient to capture large variations in appearance. Missed ground-truth is the next major error with our model. This is mainly the consequence of  exploration capacity of the sampler. Once some major object regions start providing higher scores from the score propagation, the sampler can miss other difficult instances, especially smaller objects. Thus, our sampler will not sample candidate proposals from those regions, and they remains undetected. Frequently co-occurring background regions are also challenging for the  sampler, since such regions can also accumulate higher scores over the time from the score propagation block. Those regions might also be sampled many times, resulting in detection boxes in background regions.

\subsection{Comparison with State-of-Art Methods:}
Table \ref{table:voc_sota_compare} shows a comparison of our method with state-of-the-art methods for semi- and weakly-supervised learning of object detectors. Pascal VOC 2007 is used as the fully-labeled data, while VOC 2012 is used as the unlabeled or weakly-labeled set. 
We first evaluate our method for semi-weakly supervised training.
The only other method performing Semi-weakly supervised learning is WSSOD~\cite{wssod-Fang-2021}. In this setting our method outperforms it, while being also more flexible, as it can be used with an detector model.
Compared to the model trained only on VOC07 (\emph{lower bound}), we observed a significant improvement (5.0\%) when using additional weak labeled data, approaching a model with full annotations in both datasets (\emph{upper bound}).

We then compare our model to the state-of-the-art in plain semi-supervised settings. 
To report the results of our method in the semi-supervised settings, we train a classifier on the available fully-labeled images, and used that classifier to obtain weak image-level labels for the unlabeled images. This requires training of an additional classifier, but it is less expensive than the detector pre-training used in most of the semi-supervised methods. Results in the table indicate a significant improvement in terms of performance, with the additional moderate cost of collecting weak image-level labels. 
In the semi-supervised case also, our method shows an improvement of 3.4\%, outperforming most of the methods in terms of mAP and AP 50.

\begin{table}[!h]
    \caption{Comparison in mAPs performance for different methods on VOC 2007 test set. The model is trained using VOC 2007 as fully annotated set, and VOC 2012 as the weakly annotated set. We reported the VOC style mAP (as AP 50) and COCO style mAP (as AP).}
    \centering
    \begin{tabular}{l  c  c }
        \hline
        \textbf{Method} & \textbf{AP 50} & \textbf{AP} \\
        \hline
        \hline
        \shortstack[l]{Fully Supervised \\ VOC07 \emph{(Lower Bound)}}  & 74.4 & - \\
        \hline
        \multicolumn{3}{l}{ \shortstack[l]{Semi-weakly-supervised \\ \emph {VOC07(Fully)+12(Weakly)}}} \\
        WSSOD \cite{wssod-Fang-2021}, ArXiv 2021 & 78.9  & - \\
        {Ours} & \textbf{79.4} & 47.3 \\
        \hline
        \multicolumn{3}{ l }{ \shortstack[l]{Semi-supervised \\ \emph{VOC07(Fully)+12(Unsup.)}}} \\
        CSD \cite{csd-Jeong-2019}, NeurIPS 2019 & 74.7  & 42.7 \\
        STAC \cite{stac-sohn-2015}, ArXiv 2020 & 77.4  & 44.6 \\
        WSSOD \cite{wssod-Fang-2021}, ArXiv 2021 & 78.0  & - \\
        ISD \cite{isd-jeong-2021}, CVPR 2021 & 74.4  & -\\
        {Ours}  & 77.8 & 44.2 \\
        \hline
        \shortstack[l]{Fully Supervised \\ VOC07+12 \emph{(Upper Bound)}} & 80.9 & - \\
        \hline
    \end{tabular}
    \label{table:voc_sota_compare}
\end{table}

\section{Conclusion}
\label{sec:conclusion}

This paper introduces a sampling based framework that can adapt any detector model requiring bounding boxes annotations, to weakly- and semi-supervised learning. With a single stage online learning, our method effectively makes use of the images without bounding box annotations by sampling pseudo GT boxes from the object proposals of those images based on a score accumulated for each region via a score propagation mechanism. By repeating this sampling, together with the normal detector weight update, our method is able to sample representative regions as pseudo GT boxes for the unlabeled images. With this sampling-based learning framework, training is a single stage training process. Unlike the vast majority of the semi-supervised techniques in the literature, the pseudo GT boxes are not controlled by a threshold, 
which makes the learning process easy and straightforward. Our experimental validation on VOC07 shows that our method is able to achieve excellent detection performance, with a reduced amount of fully annotated data, and additional image-level annotations.

{\small
\bibliographystyle{ieee_fullname}
\bibliography{arxiv}
}

\end{document}